\documentclass{article}

\usepackage{arxiv}

\usepackage[utf8]{inputenc} 
\usepackage[T1]{fontenc}    
\usepackage{hyperref}       
\usepackage{url}            
\usepackage{booktabs}       
\usepackage{amsfonts}       
\usepackage{nicefrac}       
\usepackage{microtype}      
\usepackage{lipsum}

\title{Offensive Language and Hate Speech Detection \\ for Danish}

\author{
  Gudbjartur Ingi Sigurbergsson\\
  Department of Computer Science\\
  IT University of Copenhagen\\
  Denmark 2300 \\
  \texttt{gusi@itu.dk} \\
   \And
  Leon Derczynski\\
  Department of Computer Science\\
  IT University of Copenhagen\\
  Denmark 2300 \\
  \texttt{ld@itu.dk} \\
}

\begin{document}
\maketitle

\begin{abstract}
The presence of offensive language on social media platforms and the implications this poses is becoming a major concern in modern society. Given the enormous amount of content created every day, automatic methods are required to detect and deal with this type of content. Until now, most of the research has focused on solving the problem for the English language, while the problem is multilingual. 


We construct a Danish dataset containing user-generated comments from \textit{Reddit} and \textit{Facebook}.
It contains user generated comments from various social media platforms, and to our knowledge, it is the first of its kind.
Our dataset is annotated to capture various types and target of offensive language.
We develop four automatic classification systems, each designed to work for both the English and the Danish language. In the detection of offensive language in English, the best performing system achieves a macro averaged F1-score of $0.74$, and the best performing system for Danish achieves a macro averaged F1-score of $0.70$. In the detection of whether or not an offensive post is targeted, the best performing system for English achieves a macro averaged F1-score of $0.62$, while the best performing system for Danish achieves a macro averaged F1-score of $0.73$. Finally, in the detection of the target type in a targeted offensive post, the best performing system for English achieves a macro averaged F1-score of $0.56$, and the best performing system for Danish achieves a macro averaged F1-score of $0.63$.

Our work for both the English and the Danish language captures the type and targets of offensive language, and present automatic methods for detecting different kinds of offensive language such as hate speech and cyberbullying. 

\end{abstract}


\section{Introduction}
\label{ch:intro}
Offensive language in user-generated content on online platforms and its implications has been gaining attention over the last couple of years. This interest is sparked by the fact that many of the online social media platforms have come under scrutiny on how this type of content should be detected and dealt with. It is, however, far from trivial to deal with this type of language directly due to the gigantic amount of user-generated content created every day. For this reason, automatic methods are required, using natural language processing (NLP) and machine learning techniques.

Given the fact that the research on offensive language detection has to a large extent been focused on the English language, we set out to explore the design of models that can successfully be used for both English and Danish. To accomplish this, an appropriate dataset must be constructed, annotated with the guidelines described in \cite{zampieri2019predicting}. We, furthermore, set out to analyze the linguistic features that prove hard to detect by analyzing the patterns that prove hard to detect.

\section{Background}
\label{ch:relatedWork}

Offensive language varies greatly, ranging from simple profanity to much more severe types of language. One of the more troublesome types of language is hate speech and the presence of hate speech on social media platforms has been shown to be in correlation with hate crimes in real life settings \cite{muller2018fanning}. It can be quite hard to distinguish between generally offensive language and hate speech as few universal definitions exist \cite{davidson2017automated}. There does, however, seem to be a general consensus that hate speech can be defined as language that targets a group with the intent to be harmful or to cause social chaos. This targeting is usually done on the basis of some characteristics such as race, color, ethnicity, gender, sexual orientation, nationality or religion \cite{schmidt2017survey}. In section \ref{sec:differentTypesOfTasks}, hate speech is defined in more detail. Offensive language, on the other hand, is a more general category containing any type of profanity or insult. Hate speech can, therefore, be classified as a subset of offensive language.  \cite{zampieri2019predicting} propose guidelines for classifying offensive language as well as the type and the target of offensive language. These guidelines capture the characteristics of generally offensive language, hate speech and other types of targeted offensive language such as cyberbullying. However, despite offensive language detection being a burgeoning field, no dataset yet exists for Danish~\cite{lacunae} despite this phenomenon being present~\cite{mm3}.

\label{sec:differentTypesOfTasks}
Many different sub-tasks have been considered in the literature on offensive and harmful language detection, ranging from the detection of general offensive language to more refined tasks such as hate speech detection \cite{davidson2017automated}, and cyberbullying detection \cite{guidelinesCyberbullying}.

A key aspect in the research of automatic classification methods for language of any kind is having substantial amount of high quality data that reflects the goal of the task at hand, and that also contains a decent amount of samples belonging to each of the classes being considered. 
To approach this problem as a supervised classification task the data needs to be annotated according to a well-defined annotation schema that clearly reflects the problem statement. The quality of the data is of vital importance, since low quality data is unlikely to provide meaningful results. 

\textbf{Cyberbullying} is  commonly defined as targeted insults or threats against an individual \cite{zampieri2019predicting}. Three factors are mentioned as indicators of cyberbullying~\cite{guidelinesCyberbullying}: intent to cause harm, repetitiveness, and an imbalance of power. This type of online harassment most commonly occurs between children and teenagers, and cyberbullying acts are prohibited by law in several countries, as well as many of the US states \cite{gregorie2001cyberstalking}.

\cite{van2015detection} focus on classifying cyberbullying events in Dutch. They define cyberbullying as textual content that is published online by an individual and is aggressive or hurtful against a victim. The annotation-schema used consists of two steps. In the first step, a three-point harmfulness score is assigned to each post as well as a category denoting the authors role (i.e. harasser, victim, or bystander). In the second step a more refined categorization is applied, by annotating the posts using the the following labels:
\textit{Threat/Blackmail}, \textit{Insult}, \textit{Curse/Exclusion}, \textit{Defamation}, \textit{Sexual Talk}, \textit{Defense}, and \textit{Encouragement to the harasser}. 

\textbf{Hate Speech}. As discussed in Section \ref{ch:intro}, hate speech is generally defined as language that is targeted towards a group, with the intend to be harmful or cause social chaos. This targeting is usually based on characteristics such as race, color, ethnicity, gender, sexual orientation, nationality or religion \cite{schmidt2017survey}. Hate speech is prohibited by law in many countries, although the definitions may vary. In article 20 of the \textit{International Covenant on Civil and Political Rights (ICCPR)} it is stated that \textit{"Any advocacy of national, racial or religious hatred that constitutes incitement to discrimination, hostility or violence shall be prohibited by law"} \cite{joseph2013international}. In Denmark, hate speech is prohibited by law, and is formally defined as public statements where a group is threatened, insulted, or degraded on the basis of characteristics such as nationality, ethnicity, religion, or sexual orientation \cite{danskHatespeechLaw}. 
Hate speech is  generally prohibited by law in the European Union, where it is defined as public incitement to violence or hatred directed against a group defined on the basis of characteristics such as race, religion, and national or ethnic origin \cite{euHateSpeechLaw}. Hate speech is, however, not prohibited by law in the United States. This is due to the fact that hate speech is protected by the freedom of speech act in the \textit{First Amendment of the U.S. Constitution} \cite{usFirstAmend}.

\cite{davidson2017automated} focus is on classifying hate speech by  distinguishing between general offensive language and hate speech. They define hate speech as \textit{"language that is used to express hatred towards a targeted group or is intended to be derogatory, to humiliate, or to insult the members of the group"}. They argue that the high use of profanity on social media makes it vitally important to be able to effectively distinguish between generally offensive language and the more severe hate speech. The dataset is constructed by gathering data from Twitter, using a hate speech lexicon to query the data with crowdsourced annotations.

\textbf{Contradicting definitions}. It becomes clear that one of the key challenges in doing meaningful research on the topic are the differences in both the annotation-schemas and the definitions used, since it makes it difficult to effectively compare results to existing work, as pointed out by several authors (\cite{abusiveLanguageDetectionInOnlineUserContent}, \cite{schmidt2017survey}, \cite{understandinAbuse}, \cite{zampieri2019predicting}). These issues become clear when comparing the work of \cite{guidelinesCyberbullying}, where racist and sexist remarks are classified as a subset of \textit{insults}, to the work of \cite{nobata2016abusive}, where similar remarks are split into two categories; \textit{hate speech} and \textit{derogatory language}. Another clear example of conflicting definitions becomes visible when comparing \cite{waseem2016hateful}, where \textit{hate speech} is considered without any consideration of overlaps with the more general type of offensive language, to \cite{davidson2017automated} where a clear distinction is made between the two, by classifying posts as either \textit{Hate speech}, \textit{Offensive} or \textit{Neither}. This lack of consensus led \cite{understandinAbuse} to propose annotation guidelines and introduce a typology.  \cite{zampieri2019semeval} argue that these proposed guidelines do not effectively capture both the type and target of the offensive language. 

\section{Dataset}
\label{sec:dataset}

In this section we give a comprehensive overview of the structure of the task and describe the dataset provided in \cite{zampieri2019predicting}. Our work adopts this framing of the offensive language phenomenon.

\subsection{Classification Structure}\label{sec:classificationStructure}

Offensive content is broken into three sub-tasks to be able to effectively identify both the type and the target of the offensive posts. These three sub-tasks are chosen with the objective of being able to capture different types of offensive language, such as hate speech and cyberbullying (section \ref{sec:differentTypesOfTasks}). 

\paragraph{Sub-task A - Offensive language identification}\label{ssec:offensEvalSubA}
In sub-task A the goal is to classify posts as either offensive or not. Offensive posts include insults and threats as well as any form of untargeted profanity \cite{zampieri2019semeval}. Each sample is annotated with one of the following labels:

\paragraph{Not Offensive (NOT).} In English this could be a post such as \textit{\#TheNunMovie was just as scary as I thought it would be. Clearly the critics don't think she is terrifyingly creepy. I like how it ties in with \#TheConjuring series}. In Danish this could be a post such as \textit{Kim Larsen var god, men hans d{\o}d blev alt for hyped}. 
\paragraph{Offensive (OFF)}. In English this could be a post such as \textit{\@USER is a \#pervert himself!}. In Danish this could be a post such as \textit{Kalle er faggot..}.

\paragraph{Sub-task B - Automatic categorization of offensive language types}\label{ssec:offensEvalSubB}
In sub-task B the goal is to classify the type of offensive language by determining if the offensive language is targeted or not. Targeted offensive language contains insults and threats to an individual, group, or others \cite{zampieri2019semeval}. Untargeted posts contain general profanity while not clearly targeting anyone \cite{zampieri2019semeval}. Only posts labeled as offensive (OFF) in sub-task A are considered in this task. Each sample is annotated with one of the following labels:
\begin{itemize}
 \setlength\itemsep{-1mm}

    \item Targeted Insult (TIN). In English this could be a post such as \textit{@USER Please ban this cheating scum}. In Danish this could be e.g. \textit{Hun skal da selv have 99 år, den smatso}. 
    \item Untargeted (UNT). In English this could be a post such as \textit{2 weeks of resp done and I still don't know shit my ass still on vacation mode}. In Danish this could e.g. \textit{Dumme svin...}
\end{itemize}

\paragraph{Sub-task C - Offensive language target identification}\label{ssec:offensEvalSubC}
In sub-task C the goal is to classify the target of the offensive language. Only posts labeled as targeted insults (TIN) in sub-task B are considered in this task \cite{zampieri2019semeval}. Samples are annotated with one of the following:
\begin{itemize}
 \setlength\itemsep{-1mm}

    \item Individual (IND): Posts targeting a named or unnamed person that is  part of the conversation. In English this could be a post such as \textit{@USER Is a FRAUD Female @USER group paid for and organized by @USER}. In Danish this could be a post such as \textit{\@USER du er sku da syg i hoved}. These examples further demonstrate that this category captures the characteristics of cyberbullying, as it is defined in section \ref{sec:differentTypesOfTasks}.
    \item Group (GRP): Posts targeting a group of people based on ethnicity, gender or sexual orientation, political affiliation, religious belief, or other characteristics. In English this could be a post such as \textit{\#Antifa are mentally unstable cowards, pretending to be relevant}. In Danish this could be e.g. \textit{Åh nej! Svensk lorteret!} 
    \item Other (OTH): The target of the offensive language does not fit the criteria of either of the previous two categories.  \cite{zampieri2019semeval}. In English this could be a post such as \textit{And these entertainment agencies just gonna have to be an ass about it.}. In Danish this could be a post such as \textit{Netto er jo et tempel over lort}.
\end{itemize}

\vspace{3mm}

One of the main concerns when it comes to collecting data for the task of offensive language detection is to find high quality sources of user-generated content that represent each class in the annotation-schema to some extent.
In our exploration phase we considered various social media platforms such as \textit{Twitter}, \textit{Facebook}, and \textit{Reddit}.\\

We consider three social media sites as data.

\begin{table}
\centering
\small
\caption{The distribution of samples by sources in our final dataset. `w off. terms" represents that the samples were retrieved using offensive words in the Danish hate speech lexicon as a filter.}
\begin{tabular}{c|c|c}
\textbf{Data Source}                       & \textbf{\# Comments} & \textbf{\% of all}\\\hline
Facebook - Ekstra Bladet               & 800                         & 22.2\\
Reddit; r/Denmark w off. term   & 200                         & 5.6\\
Reddit; r/Denmark, no off. term & 1,200                        & 33.3\\
Reddit; r/DANMAG w off. term    & 32                          & 0.9\\
Reddit; r/DANMAG                     & 1,368                       & 38.0\\
\end{tabular}
\label{table:daDatasetSources}
\end{table}

\textbf{Twitter}. Twitter has been used extensively as a source of user-generated content and it was the first source considered in our initial data collection phase. The platform provides excellent interface for developers making it easy to gather substantial amounts of data with limited efforts. However, Twitter was not a suitable source of data for our task. This is due to the fact that \textit{Twitter} has limited usage in Denmark, resulting in low quality data with many classes of interest unrepresented.

\textbf{Facebook}. We next considered \textit{Facebook}, and the public page for the Danish media company \textit{Ekstra Bladet}. 
We looked at user-generated comments on articles posted by Ekstra Bladet, and initial analysis of these comments showed great promise as they have a high degree of variation.
The user behaviour on the page and the language used ranges from neutral language to very aggressive, where some users pour out sexist, racist and generally hateful language. 
We  faced obstacles when collecting data from Facebook, due to the fact that Facebook recently made the decision to shut down all access to public pages through their developer interface. This makes computational data collection approaches impossible. 
We faced restrictions on scraping public pages with Facebook, and turned to manual collection of randomly selected user-generated comments from Ekstra Bladet's public page, yielding 800 comments of sufficient quality.

\textbf{Reddit}. Given that language classification tasks in general require substantial amounts of data, our exploration for suitable sources continued and our search next led us to \textit{Reddit}. We scraped Reddit, collecting the top 500 posts from the Danish sub-reddits \textit{r/DANMAG} and \textit{r/Denmark}, as well as the user comments contained within each post. 

We published a survey on Reddit asking Danish speaking users to suggest offensive, sexist, and racist terms for a lexicon. Language and user behaviour varies between platforms, so the goal is to capture platform-specific terms. This gave 113 offensive and hateful terms which were used to find offensive comments. The remainder of comments in the corpus were shuffled and a subset of this corpus was then used to fill the remainder of the final dataset. The resulting dataset contains 3600 user-generated comments, 800 from \textit{Ekstra Bladet on Facebook}, 1400 from \textit{r/DANMAG} and 1400 from \textit{r/Denmark}. 

In light of the \textit{General Data Protection Regulations in Europe (GDPR)} and the increased concern for online privacy, we applied some necessary pre-processing steps on our dataset to ensure the privacy of the authors of the comments that were used. Personally identifying content (such as the names of individuals, not including celebrity names) was removed. This was handled by replacing each name of an individual (i.e. author or subject) with \textit{@USER}, as presented in both \cite{zampieri2019predicting} and \cite{davidson2017automated}. All comments containing any sensitive information were removed. We classify sensitive information as any information that can be used to uniquely identify someone by the following characteristics; racial or ethnic origin, political opinions, religious or philosophical beliefs, trade union membership, genetic data, and bio-metric data.

\begin{table}
\centering
\footnotesize
\caption{The distribution of labels in the annotated Danish dataset for both the train and test set.}
\begin{tabular}{ccccc|c}
\textbf{Task A} & \textbf{Task B} & \textbf{Task C} & \textbf{Train} & \textbf{Test} & \textbf{Total} \\\hline
OFF                 & TIN                 & IND                 & 77             & 18            & 95             \\
OFF                 & TIN                 & OTH                 & 30             & 6             & 36             \\
OFF                 & TIN                 & GRP                 & 98             & 23            & 121            \\
OFF                 & UNT                 &                     & 147            & 42            & 189            \\
NOT                 &                     &                     & 2,527           & 632           & 3,159           \\\hline
\textbf{ALL}        &                     &                     & 2,879           & 721           & 3,600           \\
\end{tabular}
\label{table:daDatasetLabelDistribution}
\end{table}

We base our annotation procedure on the guidelines and schemas presented in \cite{zampieri2019predicting}, discussed in detail in section \ref{sec:classificationStructure}. As a warm-up procedure, the first 100 posts were annotated by two annotators (the author and the supervisor) and the results compared. This was used as an opportunity to refine the mutual understanding of the task at hand and to discuss the mismatches in these annotations for each sub-task.

We used a \textit{Jaccard index} \cite{jaccard} to assess the similarity of our annotations. In sub-task A the Jaccard index of these initial 100 posts was 41.9\%, 39.1\% for sub-task B , and 42.8\% for sub-task C. After some analysis of these results and the posts that we disagreed on it became obvious that to a large extent the disagreement was mainly caused by two reasons:

\begin{enumerate}
    \item Guesswork of the context where the post itself was too vague to make a decisive decision on whether it was offensive or not without more context. An example of this is a post such as \textit{Skal de hjælpes hjem, næ nej de skal sendes hjem}, where one might conclude, given the current political climate, that this is an offensive post targeted at immigrants. The context is, however, lacking so we cannot make a decisive decision. This post should, therefore, be labeled as non-offensive, since the post does not contain any profanity or a clearly stated group.
    \item Failure to label posts containing some kind of profanity as offensive (typically when the posts themselves were not aggressive, harmful, or hateful). An example could be a post like \textit{@USER sgu da ikke hans skyld at hun ikke han finde ud af at koge fucking pasta}, where the post itself is rather mild, but the presence of \textit{fucking} makes this an offensive post according to our definitions.
\end{enumerate}

In light of these findings our internal guidelines were refined so that no post should be labeled as offensive by interpreting any context that is not directly visible in the post itself  and that any post containing any form of profanity should automatically be labeled as offensive. These stricter guidelines made the annotation procedure considerably easier while ensuring consistency. The remainder of the annotation task was performed by the author, resulting in 3600 annotated samples.

\subsection{Final Dataset}\label{sec:resultingDataset}
In Table~\ref{table:daDatasetSources} the distribution of samples by sources in our final dataset is presented. Although a useful tool, using the hate speech lexicon as a filter only resulted in 232 comments. The remaining comments from Reddit were then randomly sampled from the remaining corpus.

The fully annotated dataset was split into a train and test set, while maintaining the distribution of labels from the original dataset. The training set contains 80\% of the samples, and the test set contains 20\%. Table~\ref{table:daDatasetLabelDistribution} presents the distribution of samples by label for both the train and test set. The dataset is skewed, with around $88$\% of the posts labeled as not offensive (NOT). This is, however, generally the case when it comes to user-generated content on online platforms, and any automatic detection system needs be able to handle the problem of imbalanced data in order to be truly effective.

\section{Features}

One of the most important factors to consider when it comes to automatic classification tasks the the feature representation.
This section discusses various representations used in the abusive language detection literature.  

\textbf{Top-level features.} In \cite{schmidt2017survey} information comes from top-level features such as bag-of-words, uni-grams and more complex n-grams, and the literature certainly supports this. In their work on cyberbullying detection, \cite{van2015detection} use word n-grams, character n-grams, and bag-of-words.  They report uni-gram bag-of-word features as  most predictive, followed by character tri-gram bag-of-words. Later work finds character n-grams are the most helpful features \cite{nobata2016abusive}, underlying the need for the modeling of un-normalized text. these simple top-level feature approaches are good but not without their limitations, since they often have high recall but lead to high rate of false positives \cite{davidson2017automated}. This is due to the fact that the presence of certain terms can easily lead to misclassification when using these types of features. 
Many words, however, do not clearly indicate which category the text sample belongs to, e.g. the word \textit{gay} can be used in both neutral and offensive contexts.

\textbf{Linguistic Features}
\cite{nobata2016abusive} use a number of linguistic features, including the length of samples, average word lengths, number of periods and question marks, number of capitalized letters, number of URLs, number of polite words, number of unknown words (by using an English dictionary), and number of insults and hate speech words. Although these features have not proven to provide much value on their own, they have been shown to be a good addition to the overall feature space \cite{nobata2016abusive}.

\textbf{Word Representations.} 
Top-level features often require the predictive words to occur in both the training set and the test sets, as discussed in \cite{schmidt2017survey}. For this reason, some sort of word generalization is required. 
\cite{nobata2016abusive} explore three types of embedding-derived features. First, they explore pre-trained embeddings derived from a large corpus of news samples. Secondly, they use \textit{word2vec} \cite{word2vec} to generate word embeddings using their own corpus of text samples. We use both approaches. Both the pre-trained and word2vec models represent each word as a 200 dimensional distributed real number vector. Lastly, they develop 100 dimensional \textit{comment2vec} model, based on the work of \cite{le2014distributed}. Their results show that the comment2vec and the word2vec models provide the most predictive features \cite{nobata2016abusive}. In \cite{badjatiya2017deep} they experiment with pre-trained \textit{GloVe} embeddings \cite{glove}, learned \textit{FastText} embeddings \cite{fasttext}, and randomly initialized learned embeddings. Interestingly, the randomly initialized embeddings slightly outperform the others \cite{badjatiya2017deep}.\\ 

\textbf{Sentiment Scores.}\label{ssec:semanticScores}
Sentiment scores are a common addition to the feature space of classification systems dealing with offensive and hateful speech. In our work we experiment with sentiment scores and some of our models rely on them as a dimension in their feature space. To compute these sentiment score features our systems use two Python libraries: \textit{VADER} \cite{hutto2014vader} and \textit{AFINN} \cite{nielsen2011Afinn}.Our models use the \textit{compound} attribute, which gives a normalized sum of sentiment scores over all words in the sample. The compound attribute ranges from $-1$ (extremely negative) to $+1$ (extremely positive). 

\textbf{Reading Ease.}\label{ssec:linguisticFeatures}
As well as some of the top-level features mentioned so far, we also use \textit{Flesch-Kincaid Grade Level} and \textit{Flesch Reading Ease scores}. The Flesch-Kincaid Grade Level is a metric assessing the level of reading ability required to easily understand a sample of text.

\paragraph{Pre-trained Embeddings.} The pre-trained FastText \cite{fasttext} embeddings are trained on data from the \textit{Common Crawl} project and Wikipedia, in 157 languages (including English and Danish).  FastText also provides trained models that can be used to predict word embeddings for \textit{out-of-vocabulary (OOV)} words. This is a major advantage since  challenges can arise when using pre-trained word embeddings depending on how often words in the data are not found in the pre-trained corpus. 

\textbf{Randomly Initialized Learned Embeddings.} Some of our models use randomly initialized embeddings, that are updated during training. In this case, the embedding matrix for the \textit{embedding layer} is initialized using a uniform distribution.

\section{Models}\label{sec:models}
We introduce a variety of models in our work to compare different approaches to the task at hand. First of all, we introduce naive baselines that simply classify each sample as one of the categories of interest (based on \cite{zampieri2019predicting}). Next, we introduce a logistic regression model based on the work of \cite{davidson2017automated}, using the same set of features as introduced there. Finally, we introduce three deep learning models: Learned-BiLSTM, Fast-BiLSTM, and AUX-Fast-BiLSTM. The logistic regression model is built using Scikit Learn \cite{pedregosa2011scikit} and the deep learning models are built using Keras \cite{chollet2015keras}. The following sections describe these model architectures in detail, the algorithms they are based on, and the features they use.

\paragraph{Baselines}\label{ssec:baselines}
Following the work of \cite{zampieri2019predicting}, we create simple baseline prediction models that simply classify all samples as the class containing the largest amount of samples. This allows us to investigate the properties and distribution of the samples in the datasets, and to evaluate how well our classifiers are performing. The baseline models are the following:

\paragraph{Logistic Regression}
One of our model architecture uses a \textit{Logistic Regression} as the classification algorithm. Logistic regression predicts the probability of events by using a \textit{logit} function. This logit function is usually a Sigmoid function, mapping continues variables to discrete values. A logistic regression  is computed by applying the Sigmoid function to the linear regression. Here, $y$ is the dependent variable, $X_1, \dots, X_n$ are the explanatory variables, and $\beta_0, \dots, \beta_n$ are the constants we are trying to estimate.

\begin{itemize}
 \setlength\itemsep{-1mm}
 \item Sub-Task A: All NOT for both languages.
    \item Sub-Task B: All TIN for both languages.
    \item Sub-Task C: All IND for English and All GRP for Danish. 
\end{itemize}

\paragraph{Logistic Regression Classifier}\label{lrClassifier}
We base one of our models on \cite{davidson2017automated}, where the objective is to distinguish between neutral, offensive and hateful language. 

\paragraph{Learned-BiLSTM Classifier}\label{learnedBiLSTMClassifier}
The \textit{Learned-BiLSTM model} consists of four parts; a randomly initialized embedding layer, a bi-directional long short memory (BiLSTM) layer, a fully connected hidden layer, and a fully connected output layer.  The BiLSTM layer consists of two parts; a forward and a backward LSTM, each of size 20. This vector is then used as input to the fully connected hidden layer, which contains 16 hidden units. The output is a single node for sub-tasks A and B and 3 nodes in sub-task C. The activation function used in the LSTM layers is \textit{tanh} and \textit{ReLU} is used in the hidden layer. For sub-tasks A and B, the activation function for the output layer is \textit{Sigmoid}, and \textit{Softmax} is used for sub-task C. Loss is calculated using \textit{Binary Crossentropy}.

\paragraph{Fast-BiLSTM Classifier}\label{fastBiLSTMClassifier}
The \textit{Fast-BiLSTM} model is built using the same layers and the same set of hyper-parameters as the \textit{Learned-BiLSTM} model. With this the embedding layer is initialized with the FastText embeddings. These embeddings stay fixed and are not updated during the training of the model. 

\paragraph{AUX-Fast-BiLSTM Classifier}\label{auxFastBilstmClassifier} To experiment with a wider combination of features, we extend the Fast-BiLSTM model to \textit{AUX-Fast-BiLSTM}, which accepts auxiliary features, namely: sentiment scores, n-grams weighted by their TF-IDF scores, n-gram POS-tags, counters for the number of characters, count of: syllables; words; Twitter hashtags; URLs; Twitter mentions; and re-tweets, and Flesch reading ease and grade level.

\paragraph{Hyper-Parameter Tuning}
We perform \textit{Grid Search Cross Validation}  to determine the optimal dropout amount, the batch size, the optimizer and the learning rate. The best set of hyper-parameters for all of our models are the following: batch size of 128, Adam \cite{kingma2014adam} as the optimization algorithm with a learning rate of $0.001$, and a dropout rate of $0.2$ between all layers. To tackle  imbalance in our dataset we use class weights. Each class is given a weight equal to the inverse of the number of samples it contains.

\begin{table}
\centering
\small
\caption{Results from sub-task A in English.  $\varepsilon$=epochs.}
\begin{tabular}{c|c|c}
\textbf{Model}              & \textbf{Data} & \textbf{F1$_{macro}$} \\\hline
All NOT                     & -                  & 0.419             \\
Logistic Regression         & OLID\cite{zampieri2019predicting}               & 0.724             \\
Learned-BiLSTM $\varepsilon=10$    & OLID               & 0.707             \\
\textbf{Fast-BiLSTM $\varepsilon=100$}      & \textbf{OLID}               & \textbf{0.735}             \\
AUX-Fast-BiLSTM $\varepsilon=10$   & OLID               & 0.692             \\
Logistic Regression         & OLID+HSAOFL      & 0.728             \\
Learned-BiLSTM $\varepsilon=10$  & OLID+HSAOFL      & 0.704             \\
Fast-BiLSTM $\varepsilon=100$    & OLID+HSAOFL      & 0.688             \\
AUX-Fast-BiLSTM $\varepsilon=20$ & OLID+HSAOFL      & 0.712             \\
\end{tabular}
\label{table:subAEN}
\end{table}

\begin{table}
\centering
\small
\caption{Results from  sub-task A in Danish.}\label{table:subADA}
\begin{tabular}{c|c|c}
\textbf{Model}               & \textbf{Data} & \textbf{Macro F1} \\\hline
All NOT                      & -                  & 0.467             \\
\textbf{Logistic Regression} & \textbf{DA}        & \textbf{0.699}    \\
Learned-BiLSTM (10 Epochs)   & DA                 & 0.658             \\
Fast-BiLSTM (100 Epochs)     & DA                 & 0.630             \\
AUX-Fast-BiLSTM (50 Epochs)  & DA                 & 0.675             \\
\end{tabular}
\end{table}

\begin{table*}
\small
\centering
\caption{Recall (R), precision (P), and F1 score by class for our best performing models in sub-task A.}
\begin{tabular}{c|c|c|c|c|c|c}
\textbf{Model} & \textbf{R NOT} & \textbf{R OFF} & \textbf{P NOT} & \textbf{P OFF} & \textbf{F1 NOT} & \textbf{F1 OFF} \\\hline
Fast BiLSTM EN & 0.835      & 0.646      & 0.859         & 0.603         & 0.847  & 0.624  \\
Logistic Regression DA & 0.913      & 0.506      & 0.929         & 0.450         & 0.921  & 0.476 \\
\end{tabular}
\label{table:subABest}
\end{table*}

\section{Results and Analysis}
\label{ch:results}


For each sub-task (A, B, and C, Section \ref{sec:classificationStructure}) we present results for all methods in each language.

\textbf{A - Offensive language identification:}\label{sec:resultsSubA}

\textbf{English.} For English (Table~\ref{table:subAEN}) Fast-BiLSTM performs best, trained for 100 epochs, using the OLID dataset. The model achieves a macro averaged F1-score of $0.735$. This result is comparable to the BiLSTM based methods in OffensEval. 

Additional training data from  HSAOFL \cite{davidson2017automated} does not consistently improve results. For the models using word embeddings results are worse with additional training data. 
On the other hand, for models that use a range of additional features (Logistic Regression and AUX-Fast-BiLSTM), the additional training data helps. 

\textbf{Danish.} Results are in Table~\ref{table:subADA}. Logistic Regression works best with an F1-score of $0.699$. This is the second best performing model for English, though the best performing model for English (Fast-BiLSTM) is worst for Danish.

\begin{table}
\centering
\small
\caption{Results from sub-task B in English.}\label{table:subBEN}
\begin{tabular}{c|c|c}
\textbf{Model}                      & \textbf{Data} & \textbf{Macro F1} \\\hline
All TIN                             & -                  & 0.470             \\
Logistic Regression                 & OLID               & 0.593             \\
\textbf{Learned-BiLSTM (60 Epochs)} & \textbf{OLID}      & \textbf{0.619}    \\
Fast-BiLSTM (10 Epochs)             & OLID               & 0.567             \\
AUX-Fast-BiLSTM (50 Epochs)         & OLID               & 0.595             \\
\end{tabular}
\end{table}

\begin{table}
\small
\centering
\caption{Results from  sub-task B in Danish.}\label{table:subBDA}
\begin{tabular}{c|c|c}
\textbf{Model}                        & \textbf{Data} & \textbf{Macro F1} \\\hline
All TIN                               & -                  & 0.346             \\
Logistic Regression                   & DA                 & 0.594             \\
Learned-BiLSTM (40 Epochs)            & DA                 & 0.643             \\
Fast-BiLSTM (100 Epochs)              & DA                 & 0.681             \\
\textbf{AUX-Fast-BiLSTM (100 Epochs)} & \textbf{DA}        & \textbf{0.729}    \\
\end{tabular}
\end{table}

\begin{table*}
\centering
\small
\caption{Recall (R), precision (P), and F1 score by class for our best performing models in sub-task B. 
}
\begin{tabular}{c|c|c|c|c|c|c}
\textbf{Model}                         &\textbf{ R UNT }         &\textbf{ R TIN}          & \textbf{P UNT }         & \textbf{P TIN}          & \textbf{F1 UNT}         & \textbf{F1 TIN }        \\\hline
Learned BiLSTM EN & 0.370 & 0.892 & 0.303 & 0.918 & 0.333 & 0.905 \\
AUX-Fast-BiLSTM DA& 0.690 & 0.766 & 0.725 & 0.735 & 0.707 & 0.750\\
\end{tabular}
\label{table:subBBest}
\end{table*}

\begin{table}
\small
\centering
\caption{Results for sub-task C in English.}\label{table:subCEN}
\begin{tabular}{c|c|c}
\textbf{Model}                      & \textbf{Data} & \textbf{Macro F1} \\\hline
All IND                             & -                  & 0.213             \\
Logistic Regression                 & OLID               & 0.458             \\
\textbf{Learned-BiLSTM (10 Epochs)} & \textbf{OLID}      & \textbf{0.557}    \\
Fast-BiLSTM (50 Epochs)             & OLID               & 0.516             \\
AUX-Fast-BiLSTM (40 Epochs)         & OLID               & 0.536             \\
\end{tabular}
\end{table}

\begin{table}
\small
\centering
\caption{Results from  sub-task C in Danish.}\label{table:subCDA}
\begin{tabular}{c|c|c}
\textbf{Model}                       & \textbf{Data} & \textbf{Macro F1} \\\hline
All GRP                              & -                  & 0.219             \\
Logistic Regression                  & DA                 & 0.438             \\
\textbf{Learned-BiLSTM (100 Epochs)} & \textbf{DA}        & \textbf{0.629}    \\
Fast-BiLSTM (60 epochs)              & DA                 & 0.579             \\
AUX-Fast-BiLSTM (100 Epochs)         & DA                 & 0.401             \\
\end{tabular}
\end{table}

\begin{table*}
\centering \small
\caption{Recall (R), precision (P), and F1 score by class for our best performing models in sub-task C. Baselines also included to get an idea of the class distribution.}
\begin{tabular}{c|c|c|c|c|c|c|c|c|c}
\textbf{Model}               & \textbf{R IND} & \textbf{R GRP} & \textbf{R OTH} & \textbf{P IND} & \textbf{P GRP} & \textbf{P OTH} & \textbf{F1 IND} & \textbf{F1 GRP} & \textbf{F1 OTH} \\\hline
Learned-BiLSTM EN& 0.670 & 0.667 & 0.343 & 0.770 & 0.634 & 0.273 & 0.717  & 0.650  & 0.304  \\
Learned-BiLSTM DA& 0.556 & 0.696 & 0.667 & 0.667 & 0.640 & 0.571 & 0.606  & 0.667  & 0.615 \\
\end{tabular}
\label{table:subCBest}
\end{table*}

Best results are given in Table~\ref{table:subABest}. 
The low scores for Danish compared to English may be explained by the low amount of data in the Danish dataset. The Danish training set contains $2,879$ samples (table \ref{table:daDatasetLabelDistribution}) while the English training set contains $13,240$ sample.
Futher, in the English dataset around $33\%$ of the samples are labeled offensive while in the Danish set this rate is only at around $12\%$. The effect that this under represented class has on the Danish classification task can be seen in more detail in Table~\ref{table:subABest}. 

\textbf{B - Categorization of offensive language type}\label{sec:resultsSubB}

\textbf{English.} In Table~\ref{table:subBEN} the results are presented for sub-task B on English. The Learned-BiLSTM model  trained for 60 epochs performs the best, obtaining a macro F1-score of $0.619$. 

Recall and precision scores are lower for UNT than TIN  (Table~\ref{table:subABest}). One reason is skew in the data, with only around $14\%$ of the posts labeled as UNT. The pre-trained embedding model, Fast-BiLSTM, performs the worst, with a macro averaged F1-score of $0.567$. This indicates this approach is not  good for detecting subtle differences in offensive samples in skewed data, while more complex feature models perform better.

\textbf{Danish.} Table~\ref{table:subBDA} presents the results for sub-task B and the Danish language. The best performing system is the AUX-Fast-BiLSTM model (section \ref{auxFastBilstmClassifier}) trained for 100 epochs, which obtains an impressive macro F1-score of $0.729$. This suggests that models that only rely on pre-trained word embeddings may not be optimal for this task. This is be considered alongside the indication in Section~\ref{sec:resultingDataset} that relying on lexicon-based selection also performs poorly.

The limiting factor seems to be  recall for the UNT category (Table~\ref{table:subBBest}). As mentioned in Section \ref{ch:relatedWork}, the best performing system for sub-task B in OffensEval was a rule-based system, suggesting that more refined features, (e.g. lexica) may improve  performance on this task. The better performance of models for Danish over English can most likely be explained by the fact that the training set used for Danish is more balanced, with around $42\%$ of the posts labeled as UNT. \\

\textbf{C - Offensive language target identification}\label{sec:resultsSubC}

\textbf{English.} The results for sub-task C and the English language are presented in Table~\ref{table:subCEN}. The best performing system is the Learned-BiLSTM model (section \ref{learnedBiLSTMClassifier}) trained for 10 epochs, obtaining a macro averaged F1-score of $0.557$. This is an improvement over the models introduced in \cite{zampieri2019predicting}, where the BiLSTM based model achieves a macro F1-score of $0.470$.

The main limitations of our model seems to be in the classification of OTH samples, as seen in Table~\ref{table:subCBest}. This may be explained by the imbalance in the training data. It is interesting to see that this imbalance does not effect the GRP category as much, which only constitutes about $28\%$ of the training samples. One cause for the differences in these, is the fact that the definitions of the OTH category are vague, capturing all samples that do not belong to the previous two. 

\textbf{Danish.} Table~\ref{table:subCDA} presents the results for sub-task C and the Danish language. The best performing system is the same as in English, the Learned-BiLSTM model (section \ref{learnedBiLSTMClassifier}), trained for 100 epochs, obtaining a macro averaged F1-score of $0.629$. Given that this is the same model as the one that performed the best for English, this further indicates that task specific embeddings are helpful for more refined classification tasks. 

It is interesting to see that both of the models using the additional set of features (Logistic Regression and AUX-Fast-BiLSTM) perform the worst. This indicates that these additional features are not beneficial for this more refined sub-task in Danish. The amount of samples used in training for this sub-task is very low. Imbalance does  have as much effect for Danish as it does in English, as can be seen in Table~\ref{table:subCBest}. Only about $14\%$ of the samples are labeled as OTH in the data (table \ref{table:daDatasetLabelDistribution}), but  the recall and precision scores are  closer than they are for English.

\section{Analysis}\label{sec:errorAnalysis}
We perform analysis of the misclassified samples in the evaluation of our best performing models. To accomplish this, we compute the TF-IDF scores for a range of n-grams. 
We then take the top scoring n-grams in each category and try to discover any patterns that might exist. We also perform some manual analysis of these misclassified samples. The goal of this process is to try to get a clear idea of the areas our classifiers are lacking in. The following sections describe this process for each of the sub-tasks. 

\textbf{A - Offensive language identification}

The classifier struggles to identify obfuscated offensive terms. This includes words that are concatenated together, such as \textit{barrrysoetorobullshit}. The classifier also seems to associate \textit{she} with offensiveness, and samples containing \textit{she} are misclassified as offensive in several samples while \textit{he} is less often associated with offensive language. 

There are several examples where our classifier labels profanity-bearing content as offensive that are labeled as non-offensive in the test set. Posts such as \textit{Are you fucking serious?} and \textit{Fuck I cried in this scene} are labeled non-offensive in the test set, but according to annotation guidelines should  be classified as offensive. 

The best classifier is inclined to classify longer sequences as offensive. The mean character length of misclassified offensive samples is $204.7$, while the mean character length of the samples misclassified not offensive is $107.9$. This may be due to any post containing any form of profanity being offensive in sub-task A, so more words increase the likelihood of $>0$ profane words.

The classifier suffers from the same limitations as the classifier for English when it comes to obfuscated words, misclassifying samples such as \textit{Hahhaaha lær det biiiiiaaaatch} as non-offensive. It also seems to associate the occurrence of the word \textit{svensken} with offensive language, and quite a few samples containing that word are misclassified as offensive. This can be explained by the fact that offensive language towards Swedes is common in the training data, resulting in this association. From this, we can conclude that the classifier relies too much on the presence of individual keywords, ignoring  the context of these keywords. 

\textbf{B - Categorization of offensive language type}

Obfuscation prevails in sub-task B. Our classifier misses indicators of targeted insults such as \textit{WalkAwayFromAllDemocrats}. It seems to rely too highly on the presence of profanity, misclassifying samples containing terms such as \textit{bitch, fuck, shit, etc.} as targeted insults. 

The issue of the data quality is also concerning in this sub-task, as we discover samples containing clear targeted insults such as \textit{HillaryForPrison} being labeled as untargeted in the test set. 

Our Danish classifier also seems to be missing obfuscated words such as \textit{kidsarefuckingstupid} in the classification of targeted insults. It relies to some extent to heavily on the presence of profanity such as \textit{pikfjæs, lorte} and \textit{fucking}, and misclassifies untargeted posts containing these keywords as targeted insults. 

\textbf{C - Offensive language target identification}
Misclassification based on obfuscated terms as discussed  earlier also seems to be an issue for sub-task C. This problem of obfuscated terms could be tackled by introducing character-level features such as character level n-grams.

\section{Conclusion}\label{ch:conclusion}

Offensive language on online social media platforms is harmful. Due to the vast amount of user-generated content on online platforms, automatic methods are required to detect this kind of harmful content. Until now, most of the research on the topic has focused on solving the problem for English. We explored English and Danish hate speed detection and categorization, finding that sharing information across languages and platforms leads to good models for the task.

The resources and classifiers are available from the authors under CC-BY license, pending use in a shared task; a data statement~\cite{bender-friedman-2018-data} is included in the appendix. Extended results and analysis are given in~\cite{gusi}.

\bibliographystyle{unsrt}  
\bibliography{references}

\appendix

\clearpage
\section{Data statement}
\label{appendix:datastatement}

\paragraph{Curation rationale} Examples of offensive language and hate speech, in Danish

\paragraph{Language variety} Danish, BCP-47: {\tt \small da-DK}

\paragraph{Speaker demographic} 
\begin{itemize}
    \item Danish Reddit and Facebook users
    \item Age: Unknown -- mixed.
    \item Gender: Unknown -- mixed.
    \item Race/ethnicity: Unknown -- mixed.
    \item Native language: Unknown; Danish speakers.
    \item Socioeconomic status: Unknown -- mixed.
    \item Different speakers represented: Unknown; upper bound is the number of posts.
    \item Presence of disordered speech: Some presences.
\end{itemize}

\paragraph{Annotator demographic} 
\begin{itemize}
    \item Age: 25-40. 
    \item Gender: male. 
    \item Race/ethnicity: white northern European. 
    \item Native language: Icelandic, English. 
    \item Socioeconomic status: higher education student / university faculty.
\end{itemize}

\paragraph{Speech situation} Discussions held in public on the Reddit or Facebook platform.

\paragraph{Text characteristics} Danish colloquial web speech.

\paragraph{Provenance} Originally taken from Reddit and Facebook, 2018; details given in Section~\ref{sec:dataset}.

\end{document}